\documentclass{article}
\usepackage{nips15submit_e,times}
\usepackage{url}
\usepackage[pdftex]{graphicx}
\usepackage{subfigure}
\usepackage{ctable}

\usepackage{algorithm}
\usepackage{algorithmic}
\usepackage{amsmath}
\usepackage{balance}

\usepackage{booktabs}
\usepackage{psfrag}

\newcommand{\eg}{{\it e.g.}}
\newcommand{\ie}{{\it i.e.}}

\title{DNA-Level Splice Junction Prediction using Deep Recurrent Neural Networks}

\author{
Byunghan Lee, Taehoon Lee, Byunggook Na, and Sungroh Yoon\thanks{To whom correspondence should be addressed.}\\
Electrical and Computer Engineering\\
Seoul National University\\
Seoul 08826, Korea\\
\texttt{sryoon@snu.ac.kr}\\
}

\nipsfinalcopy 

\begin{document}

\maketitle

\begin{abstract}
A eukaryotic gene consists of multiple exons (protein coding regions) and introns (non-coding regions), and a splice junction refers to the boundary between a pair of exon and intron. Precise identification of spice junctions on a gene is important for deciphering its primary structure, function, and interaction. Experimental techniques for determining exon/intron boundaries include RNA-seq, which is often accompanied by computational approaches. Canonical splicing signals are known, but computational junction prediction still remains challenging because of a large number of false positives and other complications. In this paper, we exploit deep recurrent neural networks (RNNs) to model DNA sequences and to detect splice junctions thereon. We test various RNN units and architectures including long short-term memory units, gated recurrent units, and recently proposed iRNN for in-depth design space exploration. According to our experimental results, the proposed approach significantly outperforms not only conventional machine learning-based methods but also a recent state-of-the-art deep belief network-based technique in terms of prediction accuracy.

\end{abstract}

\section{Introduction}
In eukaryotes, each gene has internal structure that consists of protein-coding regions (exons) and non-coding regions (introns). For a gene to be expressed as a protein, the nascent precursor messenger RNA (pre-mRNA) transcript is modified by splicing, which removes introns and joins exons.

Accurate identification of exon-intron boundaries (donors) and intron-exon boundaries (acceptors) from a gene sequence is important for transcriptome research (\eg, understanding alternative splicing and isoform construction~\cite{li2013truesight}) and for fully understanding the expression of the gene, given that alternative splicing affects the diversity of proteins expressed by the gene~\cite{Nilsen10}.

The location of a donor or an acceptor on a gene sequence is often called the \textit{splice junction} or \textit{splice site}. Even though the sequences near splice junctions contain common signals called \textit{canonical} splicing patterns (\eg, dimer $\mathtt{GT}$ for donors and dimer $\mathtt{AG}$ for acceptors)~\cite{Burset00}, detection of splice junctions remains challenging because of a high rate of false positives mainly caused by the short lengths of canonical splicing patterns.

We can divide existing techniques for splice junction prediction into two categories~\cite{lee2015boosted}. First, alignment-based methods~\cite{Trapnell09,Trapnell10,Wang10,Au10,Grant11} map short RNA sequences produced by RNA-seq~\cite{Chu12} to a reference genome and estimate the splicing sites, identifying the exon locations along with expression profiles. Second, machine learning (ML)-based approaches attempt to model the DNA sequences surrounding splice junctions by training with known splice site sequences. The models used by these approaches include the conventional neural networks~\cite{Stormo82,Noordewier90,Brunak91}, deep Boltzmann machines~\cite{lee2015boosted}, the support vector machine (SVM)~ \cite{Degroeve05,Huang06,Sonnenburg07}, and the hidden Markov model (HMM)~\cite{Reese97,Pertea01,Baten06}.

Recently, alignment-based techniques combined with RNA-seq are gaining popularity. However, most existing alignment-based methods, such as TopHat~\cite{Trapnell09} and SpliceMap~\cite{Au10}, consider only canonical splicing signals, often missing splicing signals of importance. For accurate junction prediction, non-canonical junction signals should be detected in addition to more salient canonical patterns. ML-based methodologies can predict non-canonical signals by appropriate training using RNA-seq data, both types of approaches can complement each other to improve the accuracy of junction prediction~\cite{li2013truesight,lee2015boosted}.

This paper presents a new ML-based approach for computational prediction of splice junctions at DNA level. Our approach relies on deep recurrent neural networks (RNNs), which are a class of artificial neural networks with feedback connections between units. Leveraged by advances in memory-based units (\eg, long short-term memory (LSTM)~\cite{hochreiter1997long} and gated recurrent unit (GRU)~\cite{cho2014properties}), architecture (\eg, stacked RNNs~\cite{pascanu2013construct} and bidirectional RNNs~\cite{schuster1997bidirectional}) and training methods (\eg, iRNN~\cite{le2015simple} and dropout~\cite{srivastava2014dropout}), RNNs are delivering breakthrough performance in tasks involving sequential modeling and prediction: \eg, speech recognition~\cite{sak2014long}, language modeling and translation~\cite{cho2014learning,bahdanau2014neural}, image caption generation~\cite{vinyals2014show}, and protein structure prediction~\cite{baldi2003principled}.

By training our RNN-based method with known splice junctions on real DNA sequences from a public sequence database, we were able to model (possibly subtle) splice signals that are otherwise difficult to learn by the existing techniques. According to our experimental results, our approach significantly outperformed not only conventional SVM-based methods but also a recently proposed state-of-the-art junction prediction technique based on deep belief networks (DBNs), in terms of prediction accuracy measured by F1-scores.




\section{Methods}\label{s:methods}

Each DNA read is a sequence four types of nucleotides and needs to be converted into numerical representations for machine learning. A widely used conversion technique is using one-hot encoding~\cite{Baldi2001,lee2015boosted}, which converts the nucleotide in each position of a DNA sequence of length $n$ into a four-dimensional binary vector and then concatenates each of the $n$ four-dimensional vectors into a $4n$-dimensional vector representing the whole sequence. 

According to our experiments, however, applying such a one-hot encoding scheme to the present problem often gives limited generalization performance caused by the sparsity of the vector representation. To overcome this limitation, our approach instead encodes each nucleotide into a four-dimensional dense vector whose elements are determined during training by the gradient descent method. The input layer of our RNN-based junction prediction method thus consists of four units.


The input layer is connected to stacked RNN layers to model DNA sequences. Any RNN unit can be used in each hidden layer, and we test three types of units in our approach: the rectified linear unit (ReLU)~\cite{le2015simple}, the LSTM unit~\cite{hochreiter1997long,gers2000learning}, and the GRU~\cite{cho2014properties}. For the LSTM unit, we use the full version that includes forget gates and peephole connections.

The outputs of the top RNN layer is fed into a fully connected output layer, which contains $K$ units for $K$-class junction prediction (\ie, $K=3$ for classifying acceptor/donor/non-site, and $K=2$ for classifying site/non-site). We use the sigmoid as the activation function for the output fully-connected layer.

For training, our approach optimizes the multi-class logarithmic loss function using Adam~\cite{kingma2014adam}, a recently proposed optimizer. In our experiments, Adam consistently gave better results than RMSprop~\cite{tieleman2012lecture}, another recent optimizer, or other conventional optimizers. When ReLUs are used in the RNN layer, we utilize the technique called iRNN~\cite{le2015simple}, which uses the identity matrix or its scaled version to initialize the recurrent weight matrix. We apply the dropout technique for regularization.

\section{Results and Discussion}
We tested our approach with the UCSC database~\cite{kent2002human} that is appropriate for three-class (\ie, acceptor/donor/non-site) classification. Among the datasets in this database, we selected the UCSC-hg38 dataset that contains 24,279 genes with 1–-173 (on average 9.44) exons per gene. For training, we randomly chose 63,454 unique exons out of the 229,255 exons in total. According to the standard procedure~\cite{noordewier1991training}, we then generated three training examples by taking the 60-mer sequences centered at the left, middle, and right boundaries of each exon. These three sequences correspond to the examples of an acceptor, a non-site, and a donor, respectively. In the same manner, we used the UCSC-hg19 dataset to generate additional training examples. From this UCSC-hg19 dataset, we randomly selected 62,819 unique exons. Note that both of the two UCSC datasets used include non-canonical splice signals in addition to canonical splice signals.

The RNN architectures tried are as follows: LSTM-based (4-60-30-3), GRU-based (4-60-30-3), and iRNN-based (4-60-3), where the first and last numbers represent the numbers of input and output layers, respectively, and the middle numbers indicate the numbers of units in the hidden RNN layers. Note that the use of the encoding scheme described in Section~\ref{s:methods}, the length of each training sample, and the number of classes justify the choice of units in the input layer, the first hidden layer, and the output layer, respectively.

\begin{figure*}
\centering
\subfigure[] {
    \includegraphics[width=0.475\linewidth]{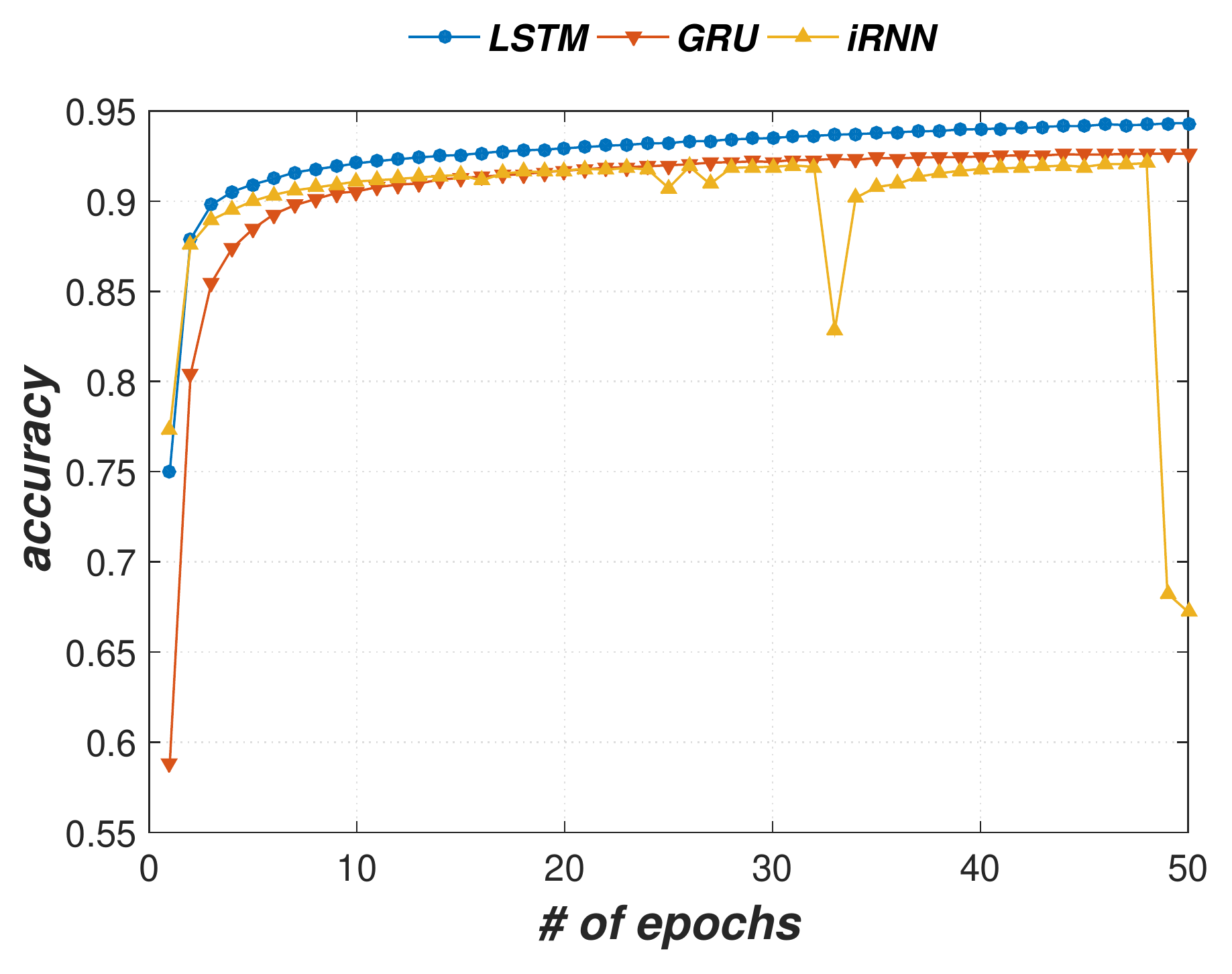}
    \label{fig:results-epoch-acc}
}
\subfigure[] {
    \includegraphics[width=0.475\linewidth]{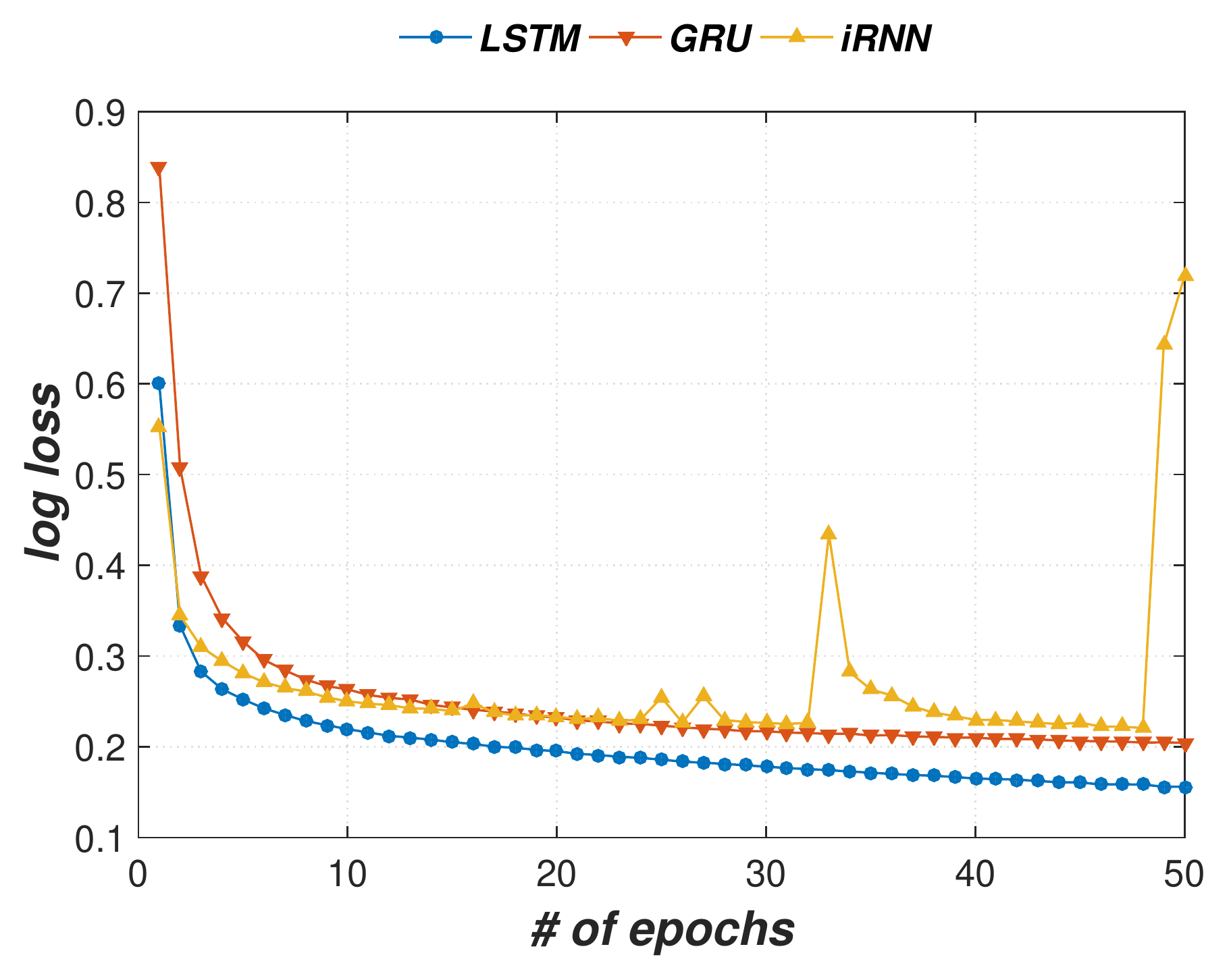}
    \label{fig:results-epoch-loss}
}
\caption{Training accuracy (a) and loss (b) of different types of building blocks of our approach.}
\end{figure*}

\ctable[
    caption = {Test accuracy of our approach and alternatives},
    label = {tbl:results-accuracy-test},
    notespar
]
{crrrr}
{}
{
    \toprule
     & \multicolumn{1}{c}{SVM (RBF)} & \multicolumn{1}{c}{SVM (sigmoid)} & \multicolumn{1}{c}{DBN~\cite{lee2015boosted}} & \multicolumn{1}{c}{Proposed (hidden unit: LSTM)} \\
    \midrule
    accuracy & 0.868 & 0.863 & 0.888 & \textbf{0.943} \\
    \bottomrule
}

Figure~\ref{fig:results-epoch-acc} and \ref{fig:results-epoch-loss} shows the training accuracy and the loss variation as the number of epochs increases. The accuracy measured in terms of F1-score\footnote{$\textrm{F1-score} = 2TP/(2TP+FP+FN)$, where $TP$, $FP$, and $FN$ represent the numbers of true positives, false positives, and false negatives, respectively.} was 0.9430, 0.9263, and 0.9210 for LSTM-based, GRU-based, and iRNN-based architectures, respectively. For each architecture, increasing the number of RNN layers further than described above did not produce significant performance gains despite substantial increases in running time. In this experiment, the LSTM-based architecture gave the best performance. The recent iRNN approach showed a fast convergence ramp-up initially (before epoch 5) but was soon outperformed by the other architectures, even showing abrupt drops and bumps for later epochs.

Table~\ref{tbl:results-accuracy-test} lists the test accuracy of the proposed junction prediction method averaged over the two UCSC datasets. For comparison, the table also includes the test accuracy of SVM (with RBF and sigmoid kernel functions) and a recently proposed junction predictor based on deep Belief networks~\cite{lee2015boosted}. Evidently, our approach outperformed not only the SVM methods but also the state-of-the-art DBN approach in terms of accuracy by large margin, giving 6.19\% increase in accuracy with respect to the DBN method.

As future work, we are planning to apply additional RNN architectures, such as bidirectional LSTM/GRU networks. Given that DNA sequences can be modeled starting from both 5' and 3' ends, considering bidirectionality is expected to boost the junction prediction performance to the next level. Another research direction would be to understand the sequence representation modeled by RNNs. This will allow us to discover novel non-canonical splice signals and eventually contribute to various transcriptome research. To this end, we first need to have a way to visualize and interpret the weight matrices of our RNN-based predictor.

\section{Conclusion}
We have described our RNN-based approach to computation prediction of splice junctions at DNA level. The proposed approach outperformed existing state-of-the-art alternatives significantly in terms of junction prediction accuracy. Given the performance of our approach, we anticipate that it will be helpful for alleviating the challenges in computational prediction of splice junctions on DNA sequences. Further, by incorporating our method into RNA-seq analysis pipelines, we expect that we can improve the performance of aligning short RNA-seq reads to a genome, more accurately identifying exon-exon splice junctions therefrom.

\bibliography{mybib}
\bibliographystyle{unsrt}

\end{document}